\documentclass[11pt,a4paper]{article}
\usepackage[hyperref]{conll-2019}
\usepackage{times}
\usepackage{latexsym}
\usepackage{fancyhdr}

\usepackage{url}

\aclfinalcopy 

\title{In Conclusion Not Repetition: Comprehensive Abstractive Summarization With Diversified Attention Based On Determinantal Point Processes}

\author{Lei Li\textsuperscript{1}, Wei Liu\textsuperscript{1}, Marina Litvak\textsuperscript{2}, Natalia Vanetik\textsuperscript{2} \and Zuying Huang\textsuperscript{1} \\
          \textsuperscript{1} Beijing University of Posts and Telecommunications \\
          \{\tt leili, thinkwee, zoehuang\}@bupt.edu.cn
          \\
          \textsuperscript{2} Shamoon College of Engineering \\
          \tt litvak.marina@gmail.com natalyav@sce.ac.il \\}

\date{}

\begin{document}
\setcounter{page}{822}
\maketitle
\begin{abstract}
Various Seq2Seq learning models designed for machine translation were applied for abstractive summarization task recently. Despite these models provide high ROUGE scores, they are limited to generate comprehensive summaries with a high level of abstraction due to its degenerated attention distribution. We introduce Diverse Convolutional Seq2Seq Model(DivCNN Seq2Seq) using Determinantal Point Processes methods(Micro DPPs and Macro DPPs) to produce attention distribution considering both quality and diversity. Without breaking the end to end architecture, DivCNN Seq2Seq achieves a higher level of comprehensiveness compared to vanilla models and strong baselines. All the reproducible codes and datasets are available online\footnote{available at https://github.com/thinkwee/DPP\_CNN\_Sum \\ marization}.
\end{abstract}

\section{Introduction}
\thispagestyle{fancy}
\definecolor{spana}{RGB}{215,47,16}
\definecolor{spanb}{RGB}{0,116,63}
\begin{table}[ht]
    \centering
    \small
        \begin{tabular}{p{7.2cm}}
        \hline
          \textbf{Article}: marseille , \textcolor{spanb}{france the french prosecutor leading an investigation into the crash of germanwings flight} 9525 insisted wednesday that \textcolor{spana}{he was not aware of any video footage} from on board the plane . marseille prosecutor brice robin told cnn that so far no videos were used in the crash investigation ...... of a cell phone video showing the harrowing final seconds from on board germanwings flight 9525 \textcolor{spanb}{as it crashed into the french alps .} ......paris match and bild reported \textcolor{spanb}{that the video was recovered from a phone at the wreckage site .} ...... cnn 's frederik pleitgen , pamela boykoff , antonia mortensen , sandrine amiel and anna-maja rappard contributed to this report . \\ \hline
          \textbf{CNN Seq2Seq}: \textcolor{spana}{french prosecutor UNK robin says he was not aware of any video .} \\ \hline
          \textbf{DivCNN Seq2Seq with Micro DPPs}: \textcolor{spanb}{new french prosecutor leading an investigation into the crash of UNK wings flight UNK 25 which crashed into french alps . the video was recovered from a phone at the wreckage site .} \\ \hline
          \textbf{DivCNN Seq2Seq with Macro DPPs}: \textcolor{spana}{french prosecutor says he was not aware of any video footage} \textcolor{spanb}{from on board UNK wings flight UNK 25 as it crashed into french alps .} \\ \hline
        \end{tabular}
    \caption{Article-summary sample from CNN-DM dataset. Colored spans are attentive parts. Micro DPPs model puts wider attention on article than vanilla does and Macro DPPs puts the widest attention, including former two models' attentive parts.}
    \label{table:sample}
\end{table}

Given an article, abstractive summarization aims at generating one or several short sentences that cover the main idea of original article, which is a combination of Natural Language Understanding(NLU) and Natural Language Generation(NLG).

Abstractive summarization uses Seq2Seq models~\cite{sutskever2014sequence} which consist of an encoder, a decoder and attention mechanism~\cite{mnih2014recurrent}. With attention mechanism the decoder can choose a weighted context representation at each generation step so it can focus on different parts of encoded information. Seq2Seq with attention achieved remarkable results on machine translation~\cite{bahdanau2014neural} and other text generation tasks such as abstractive summarizaiton~\cite{rush2015neural}. 

Unlike machine translation that emphasizes attention mechanism as a method of learning word level alignments between source text and target text, attention in summarization should be soft and diverse. Many works noticed that attention may be over concentrated for summarization and hence cause problems like generating duplicate words or duplicate sentences. Researchers try to solve these problems by introducing various attention structures, including local attention~\cite{luong2015effective}, hierarchical attention~\cite{nallapati2016abstractive}, distraction attention~\cite{chen2016distraction} and coverage mechanism~\cite{see2017get} etc. But all these works ignore another repeat problem, as we call it, "Original Text Repetition". We define and explain this problem in section 3. 

In this paper we propose a novel Diverse Convolutional Seq2Seq Model(DivCNN Seq2Seq) based on Micro Determinantal Point Processes(Micro DPPs) and Macro Determinantal Point Processes(Macro DPPs). Our contributions are as follows:
\begin{itemize}
    \item We define and describe the Original Text Repetition problem in abstractive summarization and identify the cause behind it, which are degenerated attention distributions. We have also introduced three article-related metrics for the Original Text Repetition estimation and applied them in our experiments.
    \item We suggest a solution to this problem in the form of introducing DPPs into deep neural network (DNN) attention adjustment and propose DivCNN Seq2Seq. In order to adapt DPPs to large scale computing, we propose two kinds of methods: Micro DPPs and Macro DPPs. To the best of our knowledge, this is the first attempt to adjust attention distributions considering both quality and diversity.
    \item We evaluate our models on six open datasets and show its superiority on improving the comprehensiveness of generated summaries without losing much training and inference speed.
\end{itemize}

\section{Convolutional Seq2Seq Learning}
\thispagestyle{plain}
Usually encoder and decoder in Seq2Seq architecture are recurrent neural network(RNN) or its variants like Long Short Term Memory (LSTM)~\cite{hochreiter1997long} and Gated Recurrent Unit (GRU)~\cite{chung2014empirical} network. Recently, a Seq2Seq architecture based entirely on convolutional neural networks (CNN Seq2Seq)~\cite{gehring2017convolutional} was proposed. It has better hierarchical representation of natural language and can be computed in parallel. In this paper we choose CNN Seq2Seq as our baseline system because it performs better on capturing long-term dependency, which is important for summarization.

Both encoder and decoder in CNN Seq2Seq consist of convolutional blocks. Each block contains a one dimensional convolution (Conv1d), a gated linear unit (GLU)~\cite{dauphin2017language} and several fully connected layers for dimension transformation. Residual connection~\cite{he2016deep} and batch normalization~\cite{ioffe2015batch} are used in each block. Each block receives an input $I$ of size $\mathbb{R}^{B*T*C}$, where $B$, $T$, and $C$ are respectively batch size, length of text and number of channels (the same as embedding size). Conv1d pads the sentence first and then generates a tensor $[O_1,O_2]$ of size $\mathbb{R}^{B*T*2C}$, doubling the channel. The extra channels are used in a simple non-linearity gated mechanism:
\begin{gather}
    O_1,O_2 = Conv1d(I) \\
    GLU([O_1,O_2]) = O_1 \otimes \sigma(O_2)     
\end{gather}

Multi-step attention~\cite{gehring2017convolutional} are used in CNN Seq2Seq. Each convolutional block in decoder has its own attentive context. Followed on query-key-value definition of attention, queries $Q \in \mathbb{R}^{B*T_g*C}$ are different decoder block outputs, where $T_g$ stands for summary length; keys $K \in \mathbb{R}^{B*T_s*C}$ are encoder last block outputs, where $T_s$ stands for article length; values are sum of encoder input embeddings $E \in \mathbb{R}^{B*T_s*C}$ and $K$. Because of the parallel architecture, attention for all decoder time steps can be calculated at once. Such architecture can speed up training and give convenience for our DPPs calculation. Using the simplest dot product attention, all the calculations can be done with an efficient batch matrix multiplication (BMM). 
\begin{gather}
    score_{attn} = BMM(Q,K) \\
    weight_{attn} = Softmax(score_{attn}) \\
    context = BMM(weight_{attn},K+E)
\end{gather}

\section{Original Text Repetition}
Original Text Repetition(OTR) problem means that each sentence in generated summaries are repetitions of article sentences.  The abstractive summarization hence degenerates to extractive summarization. The ROUGE metric can not detect this problem since it only measures the n-grams concurrence between generated summaries and gold summaries without taking articles into consideration. The word repeat problem~\cite{see2017get} or the lack of abstraction problem~\cite{kryscinski2018improving} can be seen as extreme condition or alternative description of OTR. Behind this phenomenon is the degenerated attention distribution learned by model which we define as: 
\begin{itemize}
    \item \textbf{Narrow Word Attention} for each summary word, the attention distribution narrows to one word position in article. 
    \item \textbf{Adjacent Sentence Attention} for all words in each summary sentence, their positions of attention peaks are adjacent or semantically adjacent, which means that attended article parts have similar features.
\end{itemize} 

\begin{figure}[t]
  \includegraphics[width=\linewidth]{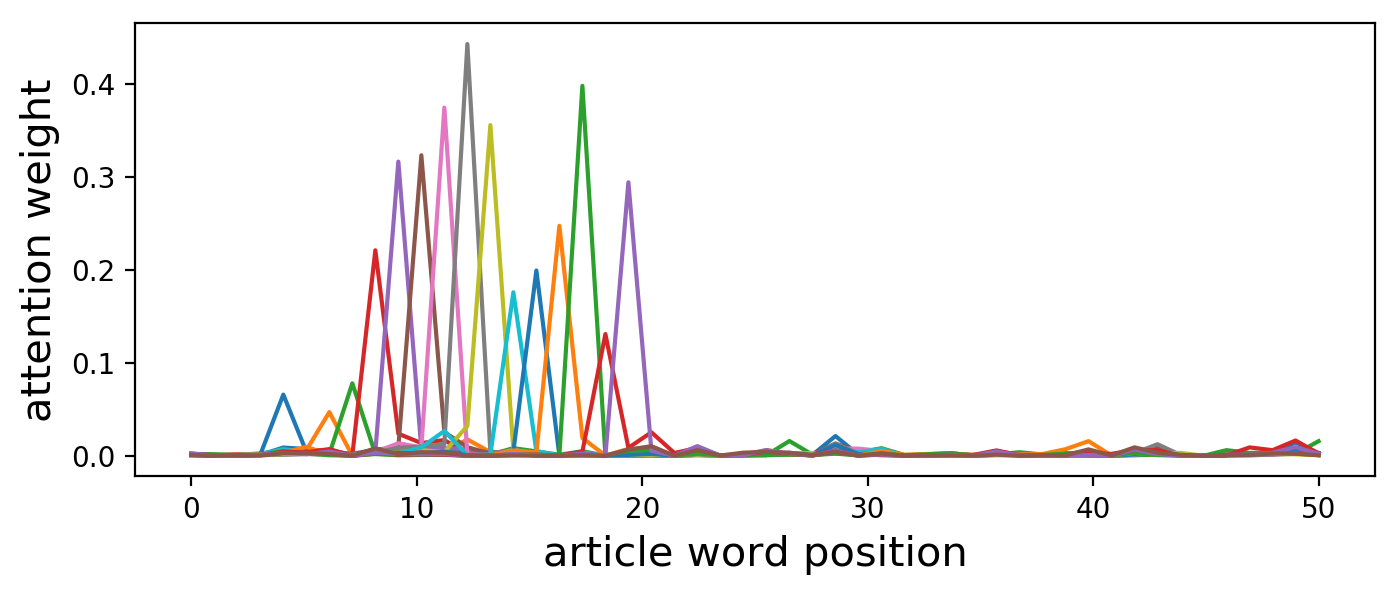}
  \caption{Degenerated attention distribution behind OTR problem. The generated summary repeats the first sentence in article. We select the first 16 words of summary and show their attention over first 50 words of article.}
  \label{fig:attention}
\end{figure}

As shown in the Figure \ref{fig:attention}, sentence attention degenerates to several adjacent peaks on the repeated article positions. Usually each sentence in gold summaries considers multiple article sentences and induct to one, not simply copying one article sentence. The gap between generated summaries(copy) and gold summaries(induce) means that model just learned to find article sentences that has the maximum similarity to gold summaries not the relation between article facts and summaries. Degenerated attention mechanism misleads the model.    
\section{Diverse Convolutional Seq2Seq Model}
\thispagestyle{plain}
To prevent Seq2Seq model from attention degeneration, we introduce DPPs as a method of regularization in CNN Seq2Seq and propose DivCNN Seq2Seq.
\subsection{Quality and Diversity Decomposition of Determinantal Point Processes}
DPPs have been widely used in recommender systems, information retrieval and extractive summarization systems. It can generate subsets with both high quality and high diversity~\cite{kulesza2011k}. 

Given a discrete, finite point process $P$ and a ground set $D$, if for every $A \in D$ and a random subset $\boldsymbol{Y}$ drawn according to P, there is:
\begin{gather}
    P(A \in \boldsymbol{Y}) = det(K_A)   
\end{gather}
where K is a real symmetric matrix that indexed by the elements of D, then P is a determinantal point process and K is the marginal kernel of DPPs. Marginal kernel merely gives marginal probability of one certain item to be selected in one particular sampling process, hence we use L-ensemble~\cite{kulesza2011k} to model atomic probabilities for every possible instantiation of Y:
\begin{gather}
    K = L(L+I)^{-1} = I - (L+I)^{-1} \\
    P_L(\boldsymbol{Y} = Y) \propto det(L_Y) \\
    P_L(\boldsymbol{Y} = Y) = \frac{det(L_Y)}{det(L+I)} \label{equ:qdscore}     
\end{gather}
L-ensemble is also one kind of DPPs and can be constructed directly using the quality($q$) and similarity($sim$) of point set:
\begin{gather}
    L_{i,j} = q(i)*sim(i,j)*q(j)  
\end{gather}
Equation \ref{equ:qdscore} is a probability that subset $Y$ being chosen, which is actually a quantitative indicator for the score of the subset considering both its quality and diversity(QD-score). Summarization follows the same principle: a good summary should consider both information significance and redundancy. In extractive summarization set of sentences with high score (quality) and diversity is chosen to a summary, using DPPs sampling algorithm~\cite{li2017uids}.

\begin{figure}[ht]
  \includegraphics[width=\linewidth]{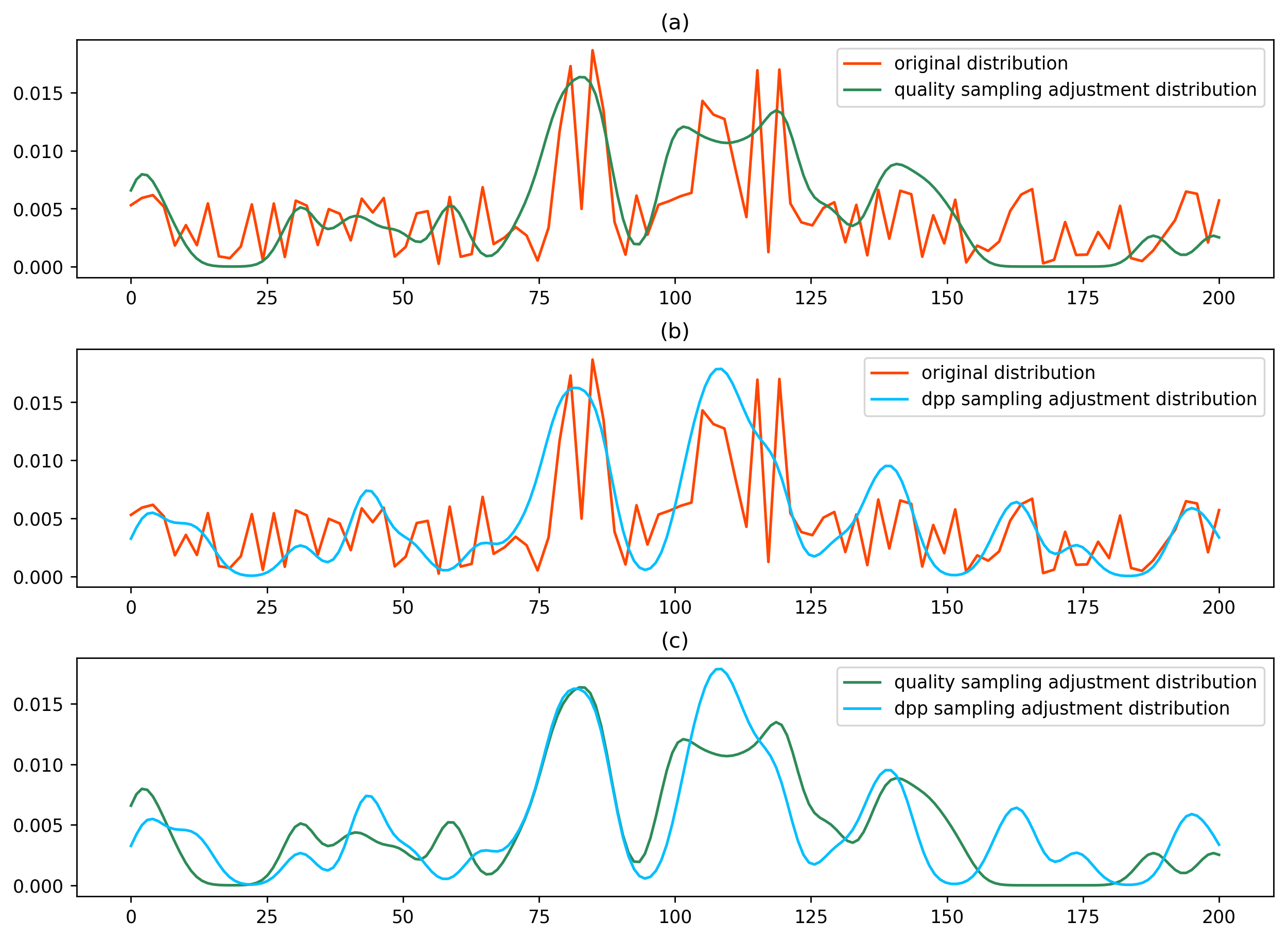}
  \caption{Comparison of different reweighting methods on a simulated distribution. DPPs sampling reweighting approximates original distribution better since it catches the high attention area around position 160. It also samples less adjacent points around position 110.}
  \label{fig:dpps}
\end{figure}

In Figure \ref{fig:dpps} we show the difference between quality-only sampling and DPPs sampling. We first generate a simulated attention distribution for testing. Then we use word position distance as similarity measure and attention as quality to construct the $L$ matrix($L$-ensemble) for DPPs. Point subset is sampled based on quality (green) or DPPs (blue), then a gaussian mixture distribution was generated around these points to soften and reweight the attention. Both samplings approximate the distribution of original attention distribution (orange), but DPPs approximate it better and have more scattering peaks. Sampling only considering attention weight (quality) generates less peaks, which means many adjacent points with low diversity are sampled.

In actual experiments we choose attention weight as quality. The model learns attention distribution to score different parts of article and obviously higher attention means higher quality. In original CNN Seq2Seq the sum of encoder output and encoder input embeddings are encoded feature vectors. We follow this setting and use the feature vectors to calculate cosine similarity. Specifically, the encoder output are tree-like semantic features extracted by CNN encoder while the encoder input embedding provides point information about a specific input element before encoding~\cite{gehring2017convolutional}. Hence feature vectors contain both highly abstract semantic features and specific grammatical features when calculating diversity. Compared to extractive summarization, DPPs in abstractive summarization use status of DNN as quality and diversity which can be optimized dynamically during training.

\begin{figure}[t]
  \includegraphics[width=\linewidth]{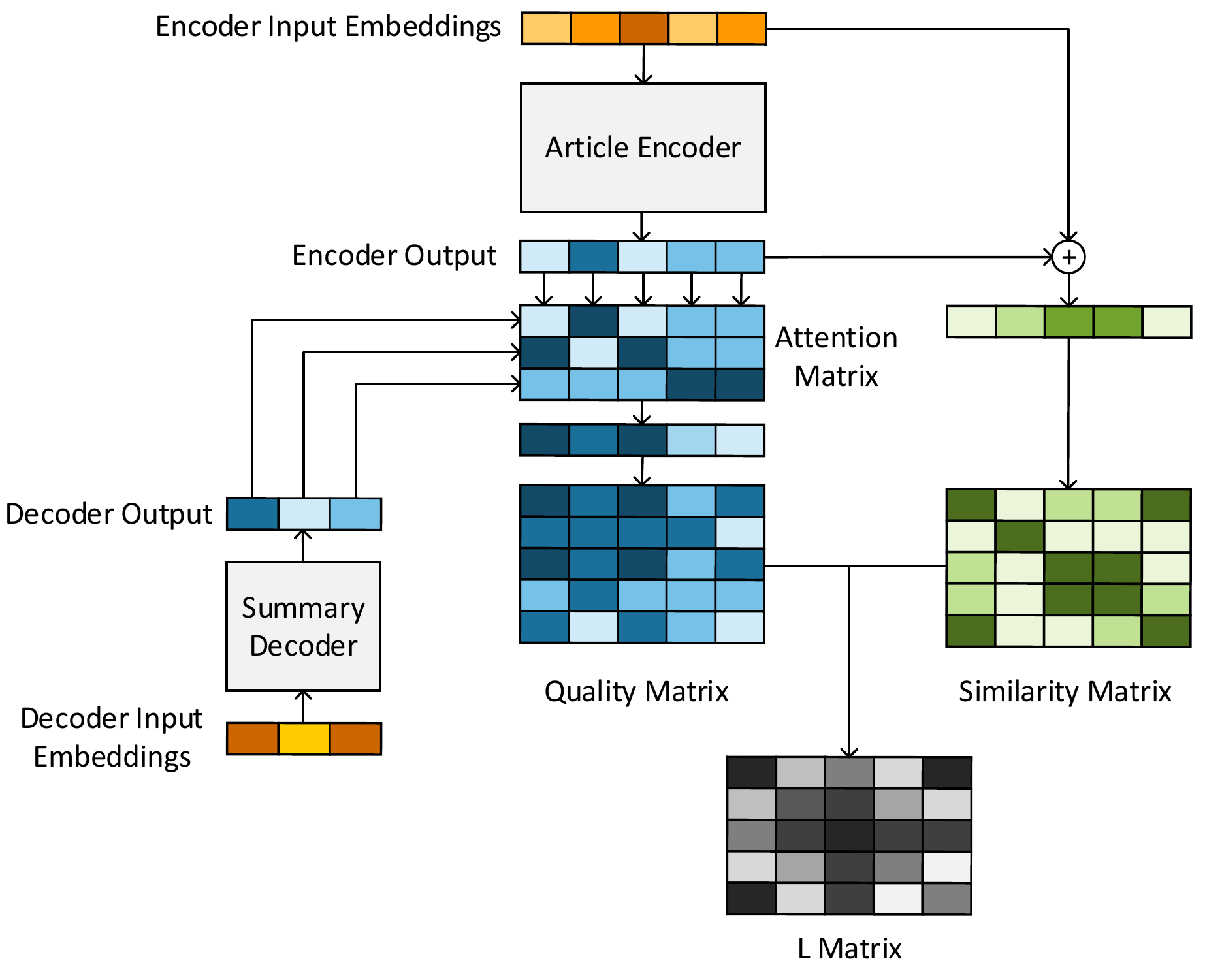}
  \caption{Construction of $L$ matrix}
  \label{fig:matrix_l}
\end{figure}

The computation of $L$ matrix is shown in Figure \ref{fig:matrix_l}. For each sample in a batch(128 in our experiments), the encoder input embeddings $E \in  \mathbb{R}^{T_s*C}$ multiply its transpose to produce similarity matrix $S \in \mathbb{R}^{T_s * T_s}$. The weight vectors of Multi-step Attention average over decoder layers and summary length, then do the same operation to generate quality matrix $Q \in R^{T_s*T_s}$. Then we use the hadamard product of $Q$ and $S$ as $L \in R^{T_s*T_s}$.

\subsection{Macro DPPs}
\thispagestyle{plain}
\begin{figure}[ht]
  \includegraphics[width=\linewidth]{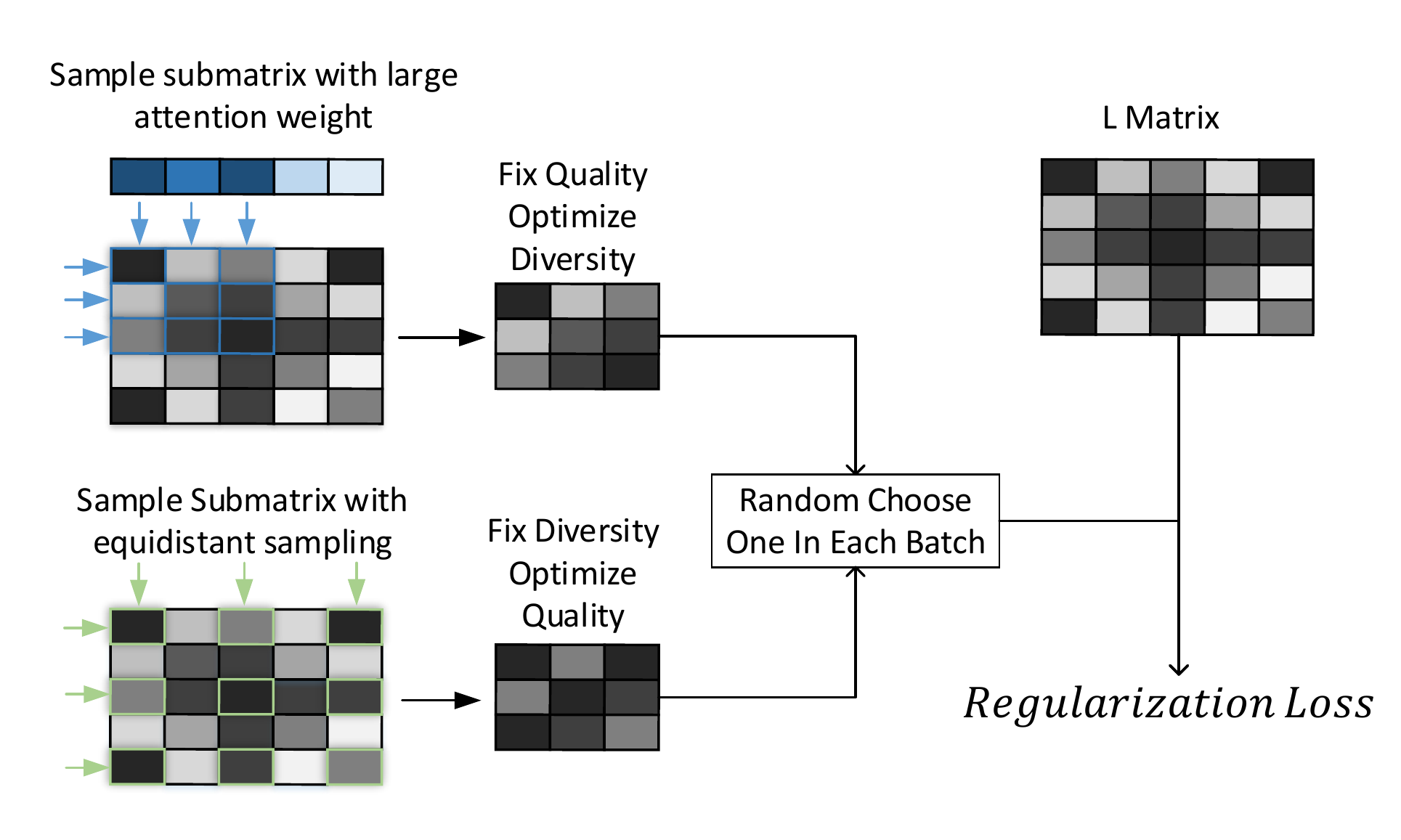}
  \caption{Conditional sampling in Macro DPPs}
  \label{fig:dpp_macro}
\end{figure}

The idea of Macro DPPs is to pick subsets under some restriction and evaluate QD-score of subset using equation \ref{equ:qdscore}. The ideal attention distribution should have subsets with high QD-score.

We do not use DPPs sampling since the purpose of Macro DPPs is to evaluate subsets not to sample subsets with high QD-score. The attention distribute over the ground set so we introduce conditional sampling to sample a subset that has high quality or high diversity, then improve the other metric as follows:
\begin{itemize}
    \item \textbf{Improve Diversity in High Quality Subset} Select points with high attention weight to construct subset and require no gradient for quality matrix, just optimize diversity.
    \item \textbf{Improve Quality in High Diversity Subset} Sampling point subset with high diversity is hard to realize, so we just make equidistant(equidistant on word positions) sampling to approximate it. Contrary to the previous method, we require no gradient for similarity matrix and just optimize quality.
\end{itemize}

We randomly choose one condition in each batch. After the point subset was chosen, the submatrix $L_Y$ can be built by selecting elements in $L$ indexed by point subset. Then we calculate the QD-score of the submatrix and add it into model loss as a regularization. We calculate the logarithmic summation of eigenvalues to prevent numerical underflow. 
\begin{gather}
    loss_{QD} = \sum \log \lambda _{L + I} - \sum \log \lambda _{L_Y}   \\
    \propto \frac{det(L_Y)}{det(L+I)} \\
    loss_{model} = \gamma loss_{MLE} + (1 - \gamma)loss_{QD}     
\end{gather}

\subsection{Micro DPPs}
\thispagestyle{plain}
The idea of Micro DPPs is to sample a subset $Y$ with large QD-score from all article positions and to use these sampled points as adjusted attention focus points. Then a Gaussian Mixture (GM) distribution around these points is generated as ideal attention distribution($weight_{ideal})$. The whole process can be seen as a selection and softening on attention. The Kullback-Leibler divergence of the ideal distribution and attention distribution ($weight$) then is added into the loss function as regularization.
\begin{gather}
    P = BFGMInference(L,t) \\
    weight_{ideal} = GM_{\mu \in P}(\mu, \sigma, \pi) \\
    loss_{KL} = KLdiv(weight_{ideal},weight_{attn}) \\
    loss_{model} = \gamma loss_{MLE} + (1 - \gamma)loss_{KL}
\end{gather}

Classic sampling algorithm for DPPs~\cite{kulesza2011k} runs slow when the size of $L$ matrix is large and it can not be computed in batch. In the DivCNN Seq2Seq model we need to construct an $L$ matrix for every sample and every layer in the decoder, which is ultimately large. To optimize DPPs runtime for this large-scale computation, we introduce a Batch computation version of Fast Greedy Maximum A Posteriori Inference~\cite{chen2018fast}(BFGMInference) to sample a subset with high QD-score.
\begin{algorithm}
\caption{BFGMInference}
\label{alg:BFGMInference}
\begin{algorithmic}[1]
\fontsize{10}{10}\selectfont
\REQUIRE matrix $L \in \mathbb{R}^{B*T_s*T_s}$, size of sampled subset $t$ 
\ENSURE Sampled subset $Y \in \mathbb{R}^{B*t}$ 
\STATE Initialize $D_i = L_{ii}$; $mask = 1^{B*T_s}$; $J = \mathop{argmax}(\log (D*mask))$; $C \in 0^{B*T_s*1}$ 
\STATE $mask_{j \in J} = 0$
\STATE $count = 1$
\WHILE{$count<t$}
    \STATE $candidate = \{ i|mask_i = 1 \}$
    \STATE $ctemp = 0^{B*T_s*1}$, $dtemp = 0^{B*T_s}$
    \FOR{$idx=0;idx<Ts-count;idx++$}
        \STATE $i = candidate[:,idx]$, $j = J$
        \STATE $e_i = (L_{j,i} - \langle c_j,c_i \rangle )/d_j$
        \STATE $ctemp_i=e_i$, $dtemp_i = {e_i}^2 $
    \ENDFOR
    \STATE $C = [C,ctemp]$, $D = D - dtemp$
    \STATE $J = \mathop{argmax}(\log (D*mask))$
    \STATE $mask_{j \in J} = 0$
    \STATE $count =  count + 1$
\ENDWHILE
\STATE $Y = \{ i|mask_i = 0 \}$
\RETURN $Y$
\end{algorithmic}
\end{algorithm}

\begin{figure}
  \includegraphics[width=\linewidth]{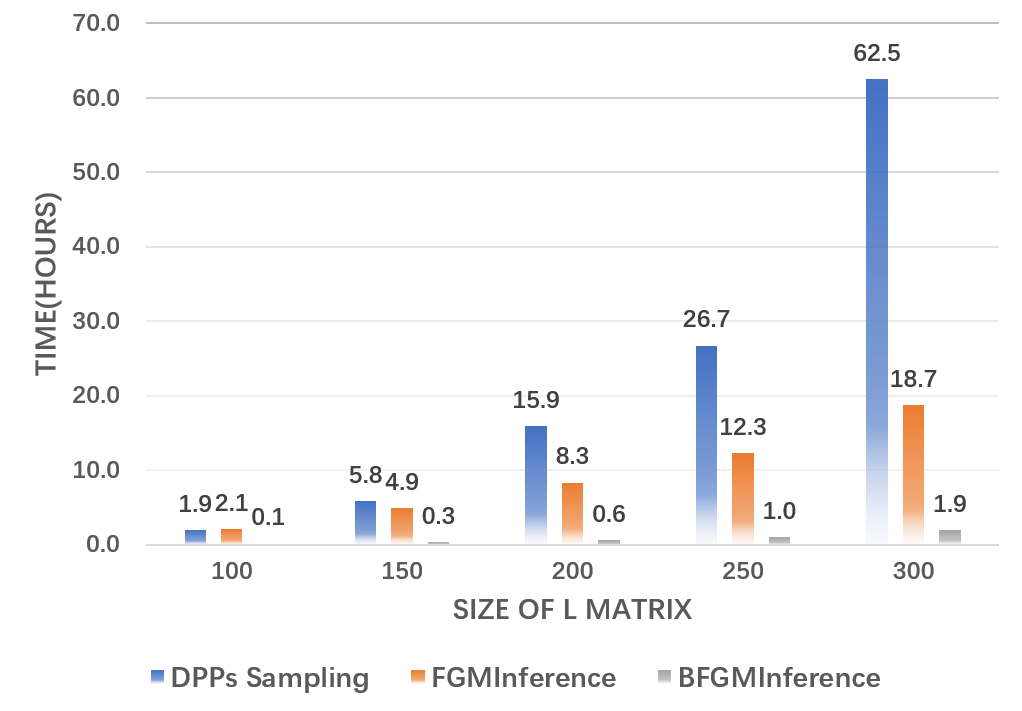}
  \caption{Speed comparison of classical DPPs sampling (blue), FGMInference (red) and BFGMInference (gray) with a batch size of 100.}
  \label{fig:speedup}
\end{figure}

BFGMInference uses a greedy method to approximate the MAP result $Y_{map} = \mathop{argmax}_{Y \in D}det(L_Y)$: each time we select $j$ that has maximum QD-score improvements and add it to $Y$.
\begin{gather}
    f(Y) = \log det(L_Y) \\
    j = \mathop{argmax}_{i \in D \backslash Y} f(Y \cup \{i\}) - f(Y) \label{equ:argmax_j}
\end{gather}
By using Cholesky decomposition we have:
\begin{gather}
    L_Y = VV^T \\
    L_{Y \cup \{i\}} = \left[ \begin{matrix} V & 0 \\ c_i & d_i \end{matrix} \right] \left[ \begin{matrix} V & 0 \\ c_i & d_i \end{matrix} \right] ^T \\
    Vc_i^T = L_{Y,i} \label{equ:c_update} \\
    d_i^2 = L_{ii} - ||c_i||_2^2 \label{equ:d_update}
\end{gather}
Then we can transform equation \ref{equ:argmax_j} into:
\begin{gather}
    j = \mathop{argmax}_{i \in D \backslash Y} \log(d_i^2)
\end{gather}
The $c$ and $d$ can be updated incrementally according to equation \ref{equ:c_update} and \ref{equ:d_update}. The complete algorithm is described in Algorithm \ref{alg:BFGMInference}. BFGMInference algorithm gains significant speed improvements when the size of $L$ matrix is large as shown in Figure \ref{fig:speedup}.

\begin{table*}[ht]
\centering
\footnotesize
\begin{tabular}{ccccc}
\hline
\textbf{Dataset}  & \textbf{\# Docs} & \textbf{Type} & \textbf{\begin{tabular}[c]{@{}c@{}}Average Document \\ Words\end{tabular}} & \textbf{\begin{tabular}[c]{@{}c@{}}Average Summary \\ Words\end{tabular}} \\ \hline
\textbf{CNN-DM}   & 287227/13368/11490          & News      & 789/777/768                                                                                    & 55/59/62                                                                                     \\ 
\textbf{NEWSROOM} & 995041/108862/108837        & News       & 659/654/652                                                                                    & 26/26/26                                                                                     \\ 
\textbf{REDDIT} & 34000/4000/4000               & Social Media & 418/445/451                                                                                    & 20/23/25                                                                                     \\ 
\textbf{BIGPATENT} & 1207222/67068/67072        & Documentation       & 699/699/699                                                                                    & 116/116/116                                                                                     \\ 
\textbf{TLDR} & 960000/20000/20000              & Social Media & 197/195/204                                                                                    & 19/19/19                                                                                     \\
\textbf{WIKIHOW} & 180000/10000/20000           & Knowledge Base    & 475/488/418                                                                                    & 62/60/74                                                                                     \\ \hline
\end{tabular}
\caption{Dataset overview(train/valid/test).}
\label{table:dataset}
\end{table*}

\section{Experimental Setup}

\thispagestyle{plain}

\textbf{Datasets} We test DivCNN Seq2Seq model on the widely used CNN-DM dataset ~\cite{hermann2015teaching} and give detailed analysis on diversity and quality. Also we tried our model on other five abstractive summarization datasets which are NEWSROOM corpus~\cite{grusky2018newsroom}, TLDR~\cite{volske2017tl}, BIGPATENT~\cite{sharma2019bigpatent}, WIKIHOW~\cite{koupaee2018wikihow} and REDDIT~\cite{kim2018abstractive}. For CNN-DM corpus we truncate articles to 600 words and summaries to 70 words. For other corpus we only keep articles and summaries that have length around its average length. Specially we only use the TIFU-long version of REDDIT and non-anonymized version of CNN-DM dataset. If the raw datasets were not divided into train/dev/test then we divide the shuffled datasets manually. Details of all six datasets are shown in Table \ref{table:dataset}. 

\thispagestyle{plain}

\textbf{Hyperparameters and Optimization} All the CNN models use a 50000 words article dictionary and 20000 words summary dictionary with byte pair encoding (BPE)~\cite{sennrich2015neural}. Word embeddings are pretrained on training corpus using Fasttext~\cite{bojanowski2017enriching,joulin2016fasttext}. We do not train models with large parameters to increase ROUGE results since what we try to improve is the comprehensiveness of each sentence in summary. The total parameters of whole CNN seq2seq Model are about 3800w and the DivCNN Seq2Seq does not change the parameters amount. All the models set embedding dimensionality and CNN channels to 256. The encoder has 20 blocks with kernel size 5 and the decoder has 4 blocks with kernel size 3. Such scale of model parameters are enough for the model to generate fluent summaries. The $\gamma$ is 0.6 for Macro DPPs and 0.7 for Micro DPPs. In Macro DPPs we choose top 30 points when optimizing diversity and a stride of 20 for equidistant. In Micro DPPs for each summary we sample 20 points to generate gaussian mixture distributions. We train the model with Nesterov's accelerated gradient method using a momentum of 0.99 and renormalized gradients when the norm exceeded 0.1~\cite{sutskever2013importance}. The beam search size is 5 and we apply a dropout of 0.1 to the embeddings and linear transform layers. We did not fix training epoches. The model was trained until the average epoch loss can not be lower anymore. The DPPs regularization only works when training and does not bring extra parameters into model. During test the model has learned proper attention so the generation is the same as vanilla CNN Seq2Seq Model.

\thispagestyle{plain}

\textbf{Article-Related Metrics} It is hard to evaluate a summary since summarization itself is very subjective. ROUGE compares generated summaries and gold summaries in checking concurrence of n-grams that results in a very limited evaluation in a word level. We set three article-related metrics to evaluate the comprehensiveness of summaries:

\thispagestyle{plain}

\begin{itemize}
    \item \textbf{Jaccard Similarity Upper Bound (JS)} For each summary sentence, we compute its jaccard similarity with every article sentence. The largest jaccard similarity for each summary sentence is selected as JS. It measures the extent to which summaries copy articles. The worst situation is 1.
    \item \textbf{Sentence Coverage (SC)} We define those article sentences that have a jaccard similarity higher than the gold summaries’ JS value as covered sentence. Then the average counts of covered article sentences for each summary sentence is a ratio that can be used to measure the coverage of the article by the summaries. The worst situation is less than or equal to 1.
    \item \textbf{Novel Bigram Proportion (NOVEL)} The percentage of bigrams in summaries that did not appear in the article. It reflects the abstraction of summaries. The worst situation is 0.
\end{itemize}  

\textbf{Strong Baseline} We chose four strong baselines which reported high ROUGE scores. BottomUp~\cite{gehrmann2018bottom}, sumGAN~\cite{liu2018generative} and RL Rerank~\cite{chen2018rewriting} are complicate systems that have additional modules or post-processings and partially relieved the OTR problem. The Pointer Generator~\cite{see2017get} reaches best ROUGE result in single end to end model but suffers greatly from repetition problem. 

\textbf{Various Possible Causes of the OTR problem} We had supposed several other reasons for the repetition problems besides attention degeneration including overfitting, bad usage of translation-style attention mechanism, lack of decoding ability and high variance attention distribution. Respectively, we designed comparative experiments as follows:
\begin{itemize}
    \item \textbf{Overfitting} We set dropout ratio to 0.1 (SMALL) and 0.5 (LARGE) for testing overfitting.
    \item \textbf{Direct Attention} Remove the encoder input embedding in attention value so the decoder looks at highly abstract features directly (DIRECT).
    \item \textbf{Lack of Decoding Ability} We double (DOUBLE) or half (HALF) the vanilla (VANILLA) decoder layers to adjust the decoding ability.
    \item \textbf{High Variance Attention} We scale down the attention distribution manually when trainning (SCALE), lowering the variance of the distribution.
\end{itemize}

\begin{table*}[ht]
\centering
\begin{tabular}{clcccccc}
\hline 
\multicolumn{2}{c}{\textbf{Model}} & \textbf{\begin{tabular}[c]{@{}c@{}}JS\end{tabular}} & \textbf{\begin{tabular}[c]{@{}c@{}}SC\end{tabular}} & \textbf{\begin{tabular}[c]{@{}c@{}}NOVEL\end{tabular}} & \textbf{\begin{tabular}[c]{@{}c@{}}R1\end{tabular}} & \textbf{\begin{tabular}[c]{@{}c@{}}R2\end{tabular}} & \textbf{\begin{tabular}[c]{@{}c@{}}RL\end{tabular}} \\ \hline
\multicolumn{2}{c}{gold} & 0.326 & -- & 0.575 & -- & -- & -- \\ \hline
\multicolumn{2}{c}{\cite{liu2018generative} sumGAN} & 0.709 & 1.136 & 0.118 & 39.92 & 17.65 & 27.25 \\
\multicolumn{2}{c}{\cite{gehrmann2018bottom} BottomUp} & \textbf{0.541} & 1.015 & 0.098 & \textbf{41.53} & \textbf{18.76} & \textbf{27.92} \\
\multicolumn{2}{c}{\cite{chen2018rewriting} RL Rerank} & 0.585 & 1.181 & 0.105 & 39.38 & 16.03 & 24.95 \\
\multicolumn{2}{c}{\cite{see2017get} Pointer Generator} & 0.774 & \textbf{1.317} & 0.079 & 39.53 & 17.28 & 26.89 \\ \hline
\multicolumn{2}{c}{CNN Seq2Seq Vanilla} & 0.616 & 1.137 & 0.167 & 30.4 & 11.7 & 23.09 \\
\multicolumn{2}{c}{DivCNN Seq2Seq with Micro DPPs} & 0.568  & 1.214  & \textbf{0.183} & 30.61 & 11.82 & 23.19 \\
\multicolumn{2}{c}{DivCNN Seq2Seq with Macro DPPs} & 0.587 & 1.265  & 0.177 & 32.28 & 12.75 & 24.32 \\ \hline
\multicolumn{2}{c}{Extract with Attention} & --  & -- & -- & 35.48 & 13.67 & 22.85 \\ 
\multicolumn{2}{c}{Extract with DPPs Diversified Attention} & -- & -- & -- & 35.35 & 13.69 & 23.07 \\ \hline
\end{tabular}
\caption{Results on CNN-DM datasets.}\label{table:cnndm_result}
\end{table*}

\section{Results}
\thispagestyle{plain}
\begin{table}[ht]
\centering
\footnotesize
\begin{tabular}{ccccc}
\hline
\multicolumn{2}{c}{\textbf{Model}} & \textbf{\begin{tabular}[c]{@{}c@{}}JS\end{tabular}} & \textbf{\begin{tabular}[c]{@{}c@{}}SC\end{tabular}} & \textbf{\begin{tabular}[c]{@{}c@{}}NOVEL\end{tabular}} \\ \hline
\multicolumn{2}{c}{SCALE} & \textbf{0.382} & 0.79 & 0.199 \\
\multicolumn{2}{c}{DIRECT} & 0.567 & 1.14 & \textbf{0.207} \\ \hline
\multicolumn{1}{c|}{\multirow{3}{*}{LARGE}} & DOUBLE & 0.591 & 1.201 & 0.192 \\
\multicolumn{1}{c|}{} & VANILLA & 0.639 & 1.261 & 0.162 \\
\multicolumn{1}{c|}{} & HALF & 0.639 & 1.281 & 0.153 \\ \hline
\multicolumn{1}{c|}{\multirow{3}{*}{SMALL}} & DOUBLE & 0.631 & 1.259 & 0.167 \\
\multicolumn{1}{c|}{} & VANILLA & 0.616 & 1.137 & 0.167 \\
\multicolumn{1}{c|}{} & HALF & 0.625 & \textbf{1.29} & 0.161 \\ \hline
\end{tabular}
\caption{Explore various possible causes of OTR.}\label{table:explore_result}
\end{table}

\textbf{Various Possibilities} As shown in Table \ref{table:explore_result}, scaled attention has the lowest jaccard similarity upper bound, which confirms our idea that over-concentrated attention makes model to copy article sentences. As for sentence coverage, small decoder with large drop out ratio performs the best, proving that large and overfitted models may have degenerated attention. Although the scaled attention has the best JS score, its SC score is the worst (SC less than 1.0 means duplicate sentences are generated). So, we may conclude that directly scaling down attention breaks the value of attention. The ideal attention is not about erasing the peak or the variance of attention but to have multiple peaks in sentence attention and have high diversity at the same time. Neither aggregative nor scattering attention distributions do good to summary generation. Direct attention model has the maximum NOVEL score which means point information about a specific input element makes model prefer copying article words instead of generating new words.

\textbf{CNN-DM Results} With large model parameters and dictionaries, four models in strong baselines reach nearly 40 points in ROUGE-1 but they perform poorly on article-related metrics. Single end to end systems like Pointer Generator performs poorly on JS value and NOVEL proportion which means most of its summaries are copied from articles. As for three models with multiple modules or post-processing, the BottomUp model has relatively good jaccard similarity upper bound and the best ROUGE result but its article-related metrics are still far away from gold summaries level. RL Rerank model has better score on JS and sumGAN has better NOVEL score but none of these model reached a balanced good performance on three article-related metrics. Compared to vanilla CNN Seq2Seq, DivCNN Seq2Seq improves the JS and NOVEL points and raises the ROUGE score at the same time, proving that proper attention distribution can help reaching a better local optima. Compared with strong baselines, DivCNN Seq2Seq achieves the best in NOVEL, second and third in JS and SC, respectively. Empirically we suggest that $\gamma$ for both Micro and Macro DPPs should be set to make average loss change less than $10\%$ compared to vanilla models. We also observed that Micro DPPs is more sensitive to $\gamma$ compared to Macro DPPs and is easier to converge but may degenerate to vanilla CNN Seq2Seq. Macro DPPs usually can reach better results but it needs more time to train since eigenvalue calculation is expensive and can not be accelerated through fp16 tensor computing. 

\textbf{Novel Bigram} NOVEL is a tricky metric which is used in many researches about abstractive summarization. There are many possibilities on explaining a high NOVEL score: first, the summary get a high Novel Bigram ratio because it has many Novel unigrams, which may be good or bad; second, the model may be underfitting and can not generate fluent sentences; third, the generated summary use novel bigrams to conclude the original text and generate readable sentences, which is the best condition. From the table \ref{table:cnndm_result} we can see that Bottom Up has the best ROUGE and JS results but worst NOVEL score. Our DivCNN Models have just the opposite metric scores. These three metrics shouldn't be ambivalent since gold summaries can reach high NOVEL and low JS at the same time. Base on these facts we make the following conjectures:
\begin{itemize}
    \item Good summaries model learned have styles differ from human-write summaries. Model tend to copy bigrams from original article and reorganize them into short summary sentences. Human tend to use brand new bigrams to paraphrase facts contained in original article. The model use a rewrite(compress and extract) way while human-write is overwrite style.
    \item Though we use human-write summaries as gold summaries for model to learn and the MLE loss is steadily descending during training but it learned summaries with different style. It implicates that model may not have a "NLU+NLG" process like human do but be restrained in a sentence-level rewrite framework. For Bottom Up and RL Rerank it is not a problem because these two systems are designed to rewrite. They only send parts of article into Seq2Seq. Such design can gain high ROUGE score but it is not the way in which human write gold summaries.
\end{itemize}

\begin{table*}[t!]
\centering
\footnotesize
\begin{tabular}{ccccc}
\hline
\textbf{Datasets} & \textbf{Baseline} & \textbf{Baseline} & \textbf{Micro DPPs} & \textbf{Macro DPPs} \\ \hline
NEWSROOM  & LEAD-3 Baseline & 30.63/21.41/28.57 & 38.33/25.01/35.3 & \textbf{40.10}/\textbf{26.71}/\textbf{37.24} \\
TLDR  & - & - & 65.94/56.97/64.65 & \textbf{66.93}/\textbf{57.84}/\textbf{65.71} \\
BIGPATENT & \cite{chen2018rewriting} & \textbf{37.12}/\textbf{11.87}/\textbf{32.45} & 33.21/10.47/24.86 & 34.55/11.65/25.96 \\
WIKIHOW  & \cite{see2017get} & \textbf{28.53}/\textbf{9.23}/\textbf{26.54} & 24.52/6.49/20.56 & 27.58/8.01/22.61 \\
REDDIT  & \cite{kim2018abstractive} & 19/3.7/15.1 & 21.39/4.24/17.11 & \textbf{21.57}/\textbf{4.48}/\textbf{17.29} \\ \hline
\end{tabular}
\caption{ROUGE F1 results(R1/R2/RL) on different datasets} \label{table:rouge_others}
\end{table*}

\begin{figure}[ht]
  \includegraphics[width=\linewidth]{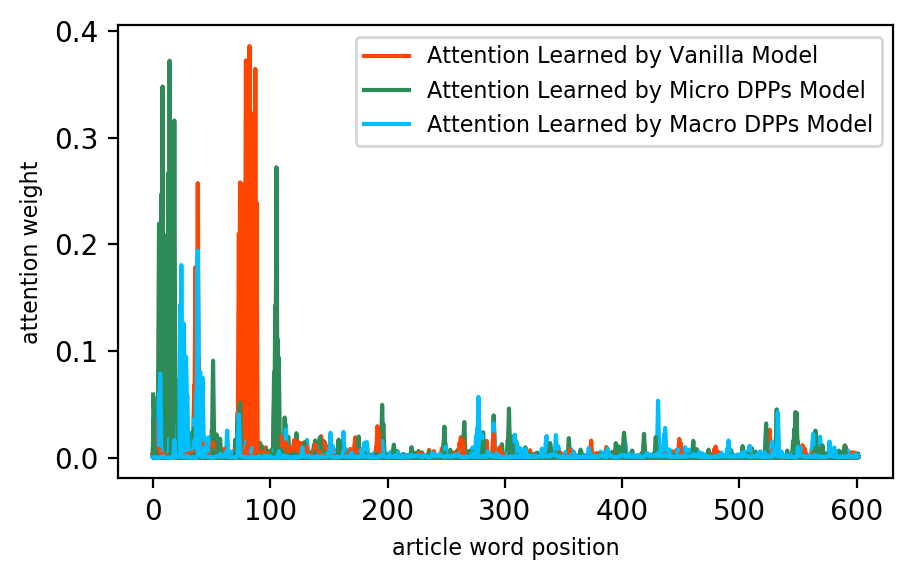}
  \caption{Actual attention distribution learned by vanilla model and DPPs models.}
  \label{fig:attn_actual}
\end{figure}
\thispagestyle{plain}
\textbf {Extractive Summarization based on Learned Attention} We also extract article sentences based on sentence attention learned by DPPs models to generate summaries. The attention of a sentence is the sum of the attention weights of the words in the sentence. Table \ref{table:cnndm_result} shows that extractive summarization reaches better ROUGE values, implicating that both vanilla and DivCNN models learned appropriate sentence attention. Extractive summarization uses accumulated sentence attention instead of specific distribution, so the results of vanilla models are almost the same as DivCNN.

\textbf{Sample Visualization} We randomly choose one sample in the test set of CNN-DM corpus to visualize and analyze. As shown in Table \ref{table:sample} we highlighted attentive parts in the article for different models. The vanilla model just generates one sentence which only focuses on one part of article. The Micro DPPs model generates two sentences considering three parts of the article. Macro DPPs considered article spans that both vanilla model  and Micro DPPs model paid attention to. We also checked the attention distribution of this sample. As shown in Figure \ref{fig:attn_actual}, vanilla model (red) learned only several peaks over article position 70 to 90, which suggests that it only focuses on one sentence and repeats this sentence in a summary. Attention learned by Micro DPPs model (green) still narrows to several peaks but explores more positions compared to vanilla. Macro DPPs (blue) has more natural design of loss function and it optimize quality and diversity directly so it has a more scattering attention distribution.

\textbf{More Datasets} We test our model on other five newly-released abstractive summarization datasets which have various compression ratio, different professional field and more flexible human-write summaries. Only ROUGE results are collected since no baseline generated summaries are provided for us to calculate article-related metrics. Table \ref{table:rouge_others} shows that DivCNN performs better than best baselines on NEWSROOM, REDDIT and reaches incredible ROUGE scores more than 60 (but no baseline is reported in the dataset paper so the result is not comparable). The compression ratio and article length have little impact on the performance of DivCNN. The results show that DivCNN prefers short summaries.

\begin{figure}[t!]
  \includegraphics[width=\linewidth]{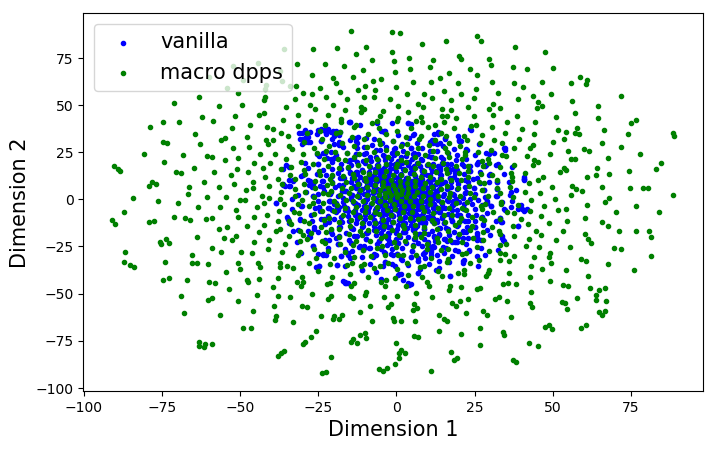}
  \caption{Presentation Degeneration Problem in NLG. We use tSNE~\cite{maaten2008visualizing} to reduce the dimension of word embeddings learned in the model.}
  \label{fig:presentation_degeneration}
\end{figure}
\thispagestyle{plain}
\textbf{Attention \& Representation Degeneration} In order to solve attention degeneration we introduce DPPs to improve the diversity of features where model paid high attention to. This solution is consistent with the Presentation Degeneration Problem in NLG~\cite{gao2018representation}. As shown in Figure \ref{fig:presentation_degeneration}, Macro DPPs have more diverse embedding presentation compared to vanilla model.~\cite{gao2018representation} directly add a regularization loss of diversity to increase the representation power of word embeddings while we aim at generating attention distribution considering both quality and diversity, resulting in learning word embeddings with rich representation power. 

\section{Conclusions and Future Works}
We have defined the "OTR" problem that leads to incomplete summaries and revealed the cause behind it, which is attention degeneration. We also introduce three article-related metrics to evaluate this problem. DPPs are applied directly on attention generation and we propose Macro and Micro DPPs versions of DivCNN Seq2Seq model to adjust attention considering both quality and diversity. Results on CNN-DM and other five open datasets show that DivCNN Seq2Seq can improve the comprehensiveness of summaries.

Due to the hardware limitation we only train a small-parameters version of DivCNN. Also we lost some precision when approximating L matrix and accelerating sampling. These drawbacks lead to limited performance improvements. In the future we hope to explore further on following directions: Quantifiable and controllable quality/diversity in DPPs; better approximation in conditional sampling, such as dynamic sampling stride adjustment; try to apply DPPs-optimized attention on another student model to improve generation.
\clearpage
\thispagestyle{plain}
\bibliography{conll-2019}

\begin{thebibliography}{33}
\expandafter\ifx\csname natexlab\endcsname\relax\def\natexlab#1{#1}\fi

\bibitem[{Bahdanau et~al.(2014)Bahdanau, Cho, and Bengio}]{bahdanau2014neural}
Dzmitry Bahdanau, Kyunghyun Cho, and Yoshua Bengio. 2014.
\newblock Neural machine translation by jointly learning to align and
  translate.
\newblock \emph{arXiv preprint arXiv:1409.0473}.

\bibitem[{Bojanowski et~al.(2017)Bojanowski, Grave, Joulin, and
  Mikolov}]{bojanowski2017enriching}
Piotr Bojanowski, Edouard Grave, Armand Joulin, and Tomas Mikolov. 2017.
\newblock Enriching word vectors with subword information.
\newblock \emph{Transactions of the Association for Computational Linguistics},
  5:135--146.

\bibitem[{Chen et~al.(2018)Chen, Zhang, and Zhou}]{chen2018fast}
Laming Chen, Guoxin Zhang, and Eric Zhou. 2018.
\newblock Fast greedy map inference for determinantal point process to improve
  recommendation diversity.
\newblock In \emph{Advances in Neural Information Processing Systems}, pages
  5622--5633.

\bibitem[{Chen et~al.(2016)Chen, Zhu, Ling, Wei, and
  Jiang}]{chen2016distraction}
Qian Chen, Xiaodan Zhu, Zhenhua Ling, Si~Wei, and Hui Jiang. 2016.
\newblock Distraction-based neural networks for document summarization.
\newblock \emph{arXiv preprint arXiv:1610.08462}.

\bibitem[{Chen and Bansal(2018)}]{chen2018rewriting}
Yen-Chun Chen and Mohit Bansal. 2018.
\newblock Fast abstractive summarization with reinforce-selected sentence
  rewriting.
\newblock \emph{arXiv preprint arXiv:1805.11080}.

\bibitem[{Chung et~al.(2014)Chung, Gulcehre, Cho, and
  Bengio}]{chung2014empirical}
Junyoung Chung, Caglar Gulcehre, KyungHyun Cho, and Yoshua Bengio. 2014.
\newblock Empirical evaluation of gated recurrent neural networks on sequence
  modeling.
\newblock \emph{arXiv preprint arXiv:1412.3555}.

\bibitem[{Dauphin et~al.(2017)Dauphin, Fan, Auli, and
  Grangier}]{dauphin2017language}
Yann~N Dauphin, Angela Fan, Michael Auli, and David Grangier. 2017.
\newblock Language modeling with gated convolutional networks.
\newblock In \emph{Proceedings of the 34th International Conference on Machine
  Learning-Volume 70}, pages 933--941. JMLR. org.

\bibitem[{Gao et~al.(2018)Gao, He, Tan, Qin, Wang, and
  Liu}]{gao2018representation}
Jun Gao, Di~He, Xu~Tan, Tao Qin, Liwei Wang, and Tieyan Liu. 2018.
\newblock Representation degeneration problem in training natural language
  generation models.

\bibitem[{Gehring et~al.(2017)Gehring, Auli, Grangier, Yarats, and
  Dauphin}]{gehring2017convolutional}
Jonas Gehring, Michael Auli, David Grangier, Denis Yarats, and Yann~N Dauphin.
  2017.
\newblock Convolutional sequence to sequence learning.
\newblock In \emph{Proceedings of the 34th International Conference on Machine
  Learning-Volume 70}, pages 1243--1252. JMLR. org.

\bibitem[{Gehrmann et~al.(2018)Gehrmann, Deng, and Rush}]{gehrmann2018bottom}
Sebastian Gehrmann, Yuntian Deng, and Alexander~M Rush. 2018.
\newblock Bottom-up abstractive summarization.
\newblock \emph{arXiv preprint arXiv:1808.10792}.

\bibitem[{Grusky et~al.(2018)Grusky, Naaman, and Artzi}]{grusky2018newsroom}
Max Grusky, Mor Naaman, and Yoav Artzi. 2018.
\newblock Newsroom: A dataset of 1.3 million summaries with diverse extractive
  strategies.
\newblock \emph{arXiv preprint arXiv:1804.11283}.

\bibitem[{He et~al.(2016)He, Zhang, Ren, and Sun}]{he2016deep}
Kaiming He, Xiangyu Zhang, Shaoqing Ren, and Jian Sun. 2016.
\newblock Deep residual learning for image recognition.
\newblock In \emph{Proceedings of the IEEE conference on computer vision and
  pattern recognition}, pages 770--778.

\bibitem[{Hermann et~al.(2015)Hermann, Kocisky, Grefenstette, Espeholt, Kay,
  Suleyman, and Blunsom}]{hermann2015teaching}
Karl~Moritz Hermann, Tomas Kocisky, Edward Grefenstette, Lasse Espeholt, Will
  Kay, Mustafa Suleyman, and Phil Blunsom. 2015.
\newblock Teaching machines to read and comprehend.
\newblock In \emph{Advances in neural information processing systems}, pages
  1693--1701.

\bibitem[{Hochreiter and Schmidhuber(1997)}]{hochreiter1997long}
Sepp Hochreiter and J{\"u}rgen Schmidhuber. 1997.
\newblock Long short-term memory.
\newblock \emph{Neural computation}, 9(8):1735--1780.

\bibitem[{Ioffe and Szegedy(2015)}]{ioffe2015batch}
Sergey Ioffe and Christian Szegedy. 2015.
\newblock Batch normalization: Accelerating deep network training by reducing
  internal covariate shift.
\newblock \emph{arXiv preprint arXiv:1502.03167}.

\bibitem[{Joulin et~al.(2016)Joulin, Grave, Bojanowski, Douze, J{\'e}gou, and
  Mikolov}]{joulin2016fasttext}
Armand Joulin, Edouard Grave, Piotr Bojanowski, Matthijs Douze, H{\'e}rve
  J{\'e}gou, and Tomas Mikolov. 2016.
\newblock Fasttext. zip: Compressing text classification models.
\newblock \emph{arXiv preprint arXiv:1612.03651}.

\bibitem[{Kim et~al.(2018)Kim, Kim, and Kim}]{kim2018abstractive}
Byeongchang Kim, Hyunwoo Kim, and Gunhee Kim. 2018.
\newblock Abstractive summarization of reddit posts with multi-level memory
  networks.
\newblock \emph{arXiv preprint arXiv:1811.00783}.

\bibitem[{Koupaee and Wang(2018)}]{koupaee2018wikihow}
Mahnaz Koupaee and William~Yang Wang. 2018.
\newblock Wikihow: A large scale text summarization dataset.
\newblock \emph{arXiv preprint arXiv:1810.09305}.

\bibitem[{Kry{\'s}ci{\'n}ski et~al.(2018)Kry{\'s}ci{\'n}ski, Paulus, Xiong, and
  Socher}]{kryscinski2018improving}
Wojciech Kry{\'s}ci{\'n}ski, Romain Paulus, Caiming Xiong, and Richard Socher.
  2018.
\newblock Improving abstraction in text summarization.
\newblock \emph{arXiv preprint arXiv:1808.07913}.

\bibitem[{Kulesza and Taskar(2011)}]{kulesza2011k}
Alex Kulesza and Ben Taskar. 2011.
\newblock k-dpps: Fixed-size determinantal point processes.
\newblock In \emph{Proceedings of the 28th International Conference on Machine
  Learning (ICML-11)}, pages 1193--1200.

\bibitem[{Li et~al.(2017)Li, Zhang, Chi, and Huang}]{li2017uids}
Lei Li, Yazhao Zhang, Junqi Chi, and Zuying Huang. 2017.
\newblock Uids: A multilingual document summarization framework based on
  summary diversity and hierarchical topics.
\newblock In \emph{Chinese Computational Linguistics and Natural Language
  Processing Based on Naturally Annotated Big Data}, pages 343--354. Springer.

\bibitem[{Liu et~al.(2018)Liu, Lu, Yang, Qu, Zhu, and Li}]{liu2018generative}
Linqing Liu, Yao Lu, Min Yang, Qiang Qu, Jia Zhu, and Hongyan Li. 2018.
\newblock Generative adversarial network for abstractive text summarization.
\newblock In \emph{Thirty-Second AAAI Conference on Artificial Intelligence}.

\bibitem[{Luong et~al.(2015)Luong, Pham, and Manning}]{luong2015effective}
Minh-Thang Luong, Hieu Pham, and Christopher~D Manning. 2015.
\newblock Effective approaches to attention-based neural machine translation.
\newblock \emph{arXiv preprint arXiv:1508.04025}.

\bibitem[{Maaten and Hinton(2008)}]{maaten2008visualizing}
Laurens van~der Maaten and Geoffrey Hinton. 2008.
\newblock Visualizing data using t-sne.
\newblock \emph{Journal of machine learning research}, 9(Nov):2579--2605.

\bibitem[{Mnih et~al.(2014)Mnih, Heess, Graves et~al.}]{mnih2014recurrent}
Volodymyr Mnih, Nicolas Heess, Alex Graves, et~al. 2014.
\newblock Recurrent models of visual attention.
\newblock In \emph{Advances in neural information processing systems}, pages
  2204--2212.

\bibitem[{Nallapati et~al.(2016)Nallapati, Zhou, Gulcehre, Xiang
  et~al.}]{nallapati2016abstractive}
Ramesh Nallapati, Bowen Zhou, Caglar Gulcehre, Bing Xiang, et~al. 2016.
\newblock Abstractive text summarization using sequence-to-sequence rnns and
  beyond.
\newblock \emph{arXiv preprint arXiv:1602.06023}.

\bibitem[{Rush et~al.(2015)Rush, Chopra, and Weston}]{rush2015neural}
Alexander~M Rush, Sumit Chopra, and Jason Weston. 2015.
\newblock A neural attention model for abstractive sentence summarization.
\newblock \emph{arXiv preprint arXiv:1509.00685}.

\bibitem[{See et~al.(2017)See, Liu, and Manning}]{see2017get}
Abigail See, Peter~J Liu, and Christopher~D Manning. 2017.
\newblock Get to the point: Summarization with pointer-generator networks.
\newblock \emph{arXiv preprint arXiv:1704.04368}.

\bibitem[{Sennrich et~al.(2015)Sennrich, Haddow, and
  Birch}]{sennrich2015neural}
Rico Sennrich, Barry Haddow, and Alexandra Birch. 2015.
\newblock Neural machine translation of rare words with subword units.
\newblock \emph{arXiv preprint arXiv:1508.07909}.

\bibitem[{Sharma et~al.(2019)Sharma, Li, and Wang}]{sharma2019bigpatent}
Eva Sharma, Chen Li, and Lu~Wang. 2019.
\newblock Bigpatent: A large-scale dataset for abstractive and coherent
  summarization.
\newblock \emph{arXiv preprint arXiv:1906.03741}.

\bibitem[{Sutskever et~al.(2013)Sutskever, Martens, Dahl, and
  Hinton}]{sutskever2013importance}
Ilya Sutskever, James Martens, George Dahl, and Geoffrey Hinton. 2013.
\newblock On the importance of initialization and momentum in deep learning.
\newblock In \emph{International conference on machine learning}, pages
  1139--1147.

\bibitem[{Sutskever et~al.(2014)Sutskever, Vinyals, and
  Le}]{sutskever2014sequence}
Ilya Sutskever, Oriol Vinyals, and Quoc~V Le. 2014.
\newblock Sequence to sequence learning with neural networks.
\newblock In \emph{Advances in neural information processing systems}, pages
  3104--3112.

\bibitem[{V{\"o}lske et~al.(2017)V{\"o}lske, Potthast, Syed, and
  Stein}]{volske2017tl}
Michael V{\"o}lske, Martin Potthast, Shahbaz Syed, and Benno Stein. 2017.
\newblock Tl; dr: Mining reddit to learn automatic summarization.
\newblock In \emph{Proceedings of the Workshop on New Frontiers in
  Summarization}, pages 59--63.

\end{thebibliography}
\thispagestyle{plain}
\bibliographystyle{acl_natbib}
\thispagestyle{plain}
\appendix

\end{document}